\documentclass[11pt,a4paper]{article}
\usepackage{times}
\usepackage{latexsym}
\usepackage{amsmath}
\usepackage{amssymb}
\usepackage{amsthm}
\usepackage[noend]{algpseudocode}
\usepackage{algorithm}
\usepackage{paralist}
\usepackage{bm}
\usepackage{url}
\usepackage{booktabs}
\usepackage{adjustbox}
\usepackage[hyperref]{conll2018}

\DeclareMathOperator*{\argmax}{arg\,max}

\newtheorem{defn}{Definition}

\newtheorem{lemma}{Lemma}

\title{A Temporally Sensitive Submodularity Framework for Timeline Summarization}

\author{Sebastian Martschat\thanks{\hspace{0.15cm}Work conducted while the author was a researcher at the Institute of Computational Linguistics, Heidelberg University.} \\
  Knowledge Architecture \& Innovation \\
  BASF SE \\
  67056 Ludwigshafen am Rhein, Germany \\
  {\tt sebastian.martschat@basf.com} \\ \And
  \hspace{1.5cm}Katja Markert \\
  \hspace{1.5cm}Institute of Computational Linguistics \\
  \hspace{1.5cm}Heidelberg University \\
  \hspace{1.5cm}69120 Heidelberg, Germany \\
  \hspace{1.5cm}{\tt markert@cl.uni-heidelberg.de} \\}

\date{}

\aclfinalcopy

\begin{document}
\maketitle
\begin{abstract}

Timeline summarization (TLS) creates an overview of long-running events via dated daily summaries for the most important dates. TLS differs from standard multi-document summarization (MDS) in the importance of date selection, interdependencies between summaries of different dates and by having very short summaries compared to the number of corpus documents. However, we show that MDS optimization models using submodular functions can be adapted to yield well-performing TLS models by designing objective functions and constraints that model the temporal dimension inherent in TLS. Importantly, these adaptations retain the elegance and advantages of the original MDS models (clear separation of features and inference, performance guarantees and scalability, little need for supervision) that current TLS-specific models lack. An open-source implementation of the framework and all models described in this paper is available online.\footnote{\url{http://smartschat.de/software}}

\end{abstract}


\section{Introduction}
\label{sec:intro}

There is an abundance of reports on events, crises and
disasters. \emph{Timelines} (see Table~\ref{table:example}) summarize and
date these reports in an ordered overview.  \emph{Automatic
  Timeline Summarization} (TLS)
constructs such timelines from corpora that contain  articles
about the corresponding event.

In contrast to standard \emph{multi-document summarization} (MDS), in
TLS we need to explicitly model the temporal dimension of the task,
specifically we need to select the most important dates for a long-running event
and summarize each of these dates. In addition, TLS 
deals with a much larger number of
documents to summarize, enhancing scalability and redundancy
problems.  These differences have significant consequences for
constraints, objectives, compression rates and scalability (see
Section~\ref{subsec:task-mds}).

Due to these differences, most
work on TLS has been separate from the MDS community.\footnote{The TLS
  systems in \cite{yanrui11a,tran13b} are compared to some simple MDS
  systems as baselines, but not to state-of-the art ones.}  Instead,
approaches to TLS start from scratch, optimizing task-specific
heuristic criteria \citep[inter
  alia]{chieu04,yanrui11a,wangwilliam16}, often with manually
determined parameters \cite{chieu04,yanrui11a} or needing supervision
\cite{wangwilliam16}. 
As features and architectures  are
rarely reused or indeed separated from each other,
it is difficult to assess reported
improvements. Moreover, none of these approaches give performance guarantees for the task, which are possible in MDS models
 based on function optimization \citep{mcdonald07,linhui11} that yield state-of-the art models for MDS \cite{hongkai14b,hirao17}.

\begin{table}[t]\centering
  \footnotesize
  \begin{tabular}{@{}p{7.5cm}@{}}
  \toprule
  \textbf{2011-03-16}\\
  Security forces break up a gathering in Marjeh Square in Damascus of 150 protesters holding pictures of imprisoned relatives. Witnesses say 30 people are arrested.\\
  \textbf{2011-03-24}\\
  President Bashar al-Assad orders the formation of a committee to study how to raise living standards and lift the law covering emergency rule, in place for 48 years.\\
  \textbf{2011-03-29}\\
  Government resigns.\\
  \bottomrule
  \end{tabular}
  \caption{\label{table:example}Excerpt from a Syrian War Reuters timeline.}
\end{table}


In this paper we take a step back from the differences between MDS and
TLS and consider the following question: \emph{Can MDS optimization models be
  expanded to yield scalable, well-performing TLS models that take
  into account the  temporal properties of TLS, while keeping
  MDS advantages such as modularity and performance guarantees?}
In particular, we make the following contributions:




\begin{compactitem}

\item We adapt the submodular function model of \citet{linhui11} to
  TLS (Section~\ref{sec:submod}). This framework is scalable and
  modular, allowing a ``plug-and-play'' approach for different
  submodular functions.  It needs  little supervision or parameter
  tuning.  We show that even this straightforward MDS adaptation
  equals or outperforms two strong TLS baselines on two corpora
  for most metrics.

\item We modify the MDS-based objective function  by adding temporal criteria
  that take date selection and interdependencies between daily summaries into account
    (Section~\ref{sec:obj}).

\item We then add more complex temporal
  constraints,  going
  beyond the simple cardinality constraints in MDS
  (Section~\ref{sec:constraints}). These new
  constraints specify the uniformity of the timeline daily summaries
  and date distribution. We also give the first 
  performance guarantees for TLS using  these  constraints.

\item We propose a TLS evaluation framework, in which we study the
  effect of temporal objective functions and constraints. We show
  performance improvements of our temporalizations
  (Section~\ref{sec::exps}).  We also present the first oracle
  upper bounds for the problem and study the impact that timeline
  properties, such as compression rates, have on performance.



\end{compactitem}


\section{Timeline Summarization} 
\label{sec:task}

Given a query (such as \textit{Syrian war}) TLS needs to (i) extract
the most important events for the query and their corresponding dates
and (ii) obtain concise daily summaries for each selected date
\cite{allan01,chieu04,yanrui11a,tran15b,wangwilliam16}.

\subsection{Task Definition and Notation}

A \emph{timeline} is a sequence $(d_1,v_1),\ldots,(d_k,v_k)$ where the
$d_i$ are dates and the $v_i$ are summaries for the dates $d_i$. Given are a
query $q$ and an associated corpus $C$ that contains documents relevant to the
query. The task of \emph{timeline summarization} is to generate a timeline $t$
based on $C$. The number of dates in $t$ as well as the length of the daily summaries are typically controlled by the
user.

We denote with $U$ the set of sentences in $C$. We assume that each sentence
in $U$ is dated (either by a date expression appearing in the sentence or by
the publication date of the article it appears in). For a sentence $s$ we write
$d(s)$ for the date of $s$.

\subsection{Relation to MDS} 
\label{subsec:task-mds}

In MDS, we also need to generate a (length-limited) summary of texts in a 
corpus $C$ (with an optional query $q$ used
to retrieve the corpus). In the traditional DUC multi-document summarization tasks\footnote{\url{https://duc.nist.gov/}},
most tasks are either not event-based at all or concentrate on one single event.
In contrast, in TLS, the corpus
describes an event that consists of several subevents that happen on
different days.

This difference has substantial effects. In MDS, criteria (such as
coverage and diversity) and length constraints 
apply on a global level. In TLS, the whole summary is naturally
divided into per-day summaries. Criteria and
constraints apply on a global level as well as on a per-day
level.

Even for the small number of DUC tasks that do focus on longer-running
events, several differences to TLS still hold.  First, the
temporal dimension in the DUC gold standard summaries and system outputs
is playing a minor role, with few explicit datings of events and a
non-temporal structure of the output, leading again to the
above-mentioned differences in constraints and criteria.  The ROUGE
evaluation measures used in MDS \cite{linchinyew04} also do not take into account temporality and do
not explicitly penalize wrong datings. Second, corpora in TLS
typically contain thousands of documents per query
\citep{tran13a,tran15b}. This is magnitudes larger than the corpora
usually considered for MDS \citep{over04}. This leads to a low
compression rate\footnote{\emph{Compression rate} is the length of the
  summary divided by the length of the source \citep{radev04a}.} and
requires approaches to be scalable.


\section{Casting TLS as MDS}
\label{sec:submod}

In the introduction, we identified several issues in existing TLS
research, including lack of modularity, insufficient separation
between features and model, and the lack of performance
guarantees. Global constrained optimization frameworks used in MDS
\citep{mcdonald07,linhui11} do separate constraints, features
and inference and allow for
optimal solutions or solutions with performance guarantees. They also can be used in an unsupervised manner. We now
cast TLS as MDS, employing constraints and criteria used
for standard MDS \cite{linhui11}. 
While this ignores the temporal
dimension of TLS, it will give us a baseline
and a starting point for systematically
incorporating temporal information.

\subsection{Problem Statement and Inference}

We can understand summarization as an optimization of an objective function
that evaluates sets of sentences over constraints.
Hence, let $U$ be a
set of sentences in a corpus and let $f\colon\, 2^U \to
\mathbb{R}_{\ge 0}$ be a function that measures the quality of a
summary. Let $\mathcal{I} \subseteq \{X ~ \vert ~ X \in 2^U\}$ be a
set of constraints\footnote{An example are length constraints, which
  can be expressed as $\mathcal{I} = \{X ~ \vert ~ \left|X\right| \le
  m, X \in 2^U\}$ for some $m$.}. We then consider the optimization
problem
\begin{equation}
  \label{eqn::problem}
  S^* = \argmax_{S \subseteq U, S \in \mathcal{I}} f(S).
\end{equation}


Solving Equation \ref{eqn::problem} exactly
does not scale well \citep{mcdonald07} and is therefore inappropriate
for the large-scale data used in TLS.  The greedy
Algorithm~\ref{alg:greedy} that iteratively constructs an output
solves the equation approximately (also used in \citet{mcdonald07} and
\citet{linhui11}).
 

\begin{algorithm}
\begin{algorithmic}
\Require{A set of sentences $U$, a function $f$, a set of constraints $\mathcal{I}$}
\Function{Greedy}{$U$, $f$, $\mathcal{I}$}
    \State{Set $S = \emptyset$, $K = U$}
    \While{$K \neq \emptyset$}
      \State{$s = \argmax_{t \in K} f(S \cup \{t\}) - f(S)$}
      \If{$S \cup \{s\} \in \mathcal{I}$}
        \State{$S = S \cup \{s\}$}
      \EndIf
      \State{$K = K \setminus \{s\}$}
    \EndWhile
\EndFunction
\Ensure{A summary $S$}
\end{algorithmic}
\caption{\label{alg:greedy}Greedy algorithm.}
\end{algorithm}

\subsection{Monotonicity and Submodularity}

The results obtained by \textsc{Greedy} can be
arbitrarily bad. However,  there
are performance guarantees if the objective function $f$ and the
constraints $\mathcal{I}$ are ``sufficiently nice''
\citep{calinescu11}. Many results rely on objective functions
that are \emph{monotone} and \emph{submodular}. A function $f$ is
monotone if $A \subseteq B$ implies that $f(A) \le f(B)$. A function
$f$ is submodular if it possesses a ``diminishing returns property'',
i.e.\ if for $A \subseteq B \subset U$ and $v \in U \setminus B$ we
have $f(A \cup \{v\}) - f(A) \ge f(B \cup \{v\}) - f(B)$.


From now on we assume that the function $f$ is of the form $f \equiv \sum_{i=1}^m f_i$ with monotone submodular $f_i\colon\, U \to \left[0,1\right]$ ($i \in \{1,\ldots,m\}$). We normalize all $f_i$ to $\left[0,1\right]$. By closure properties of monotonicity and submodularity, $f$ is also submodular.


\subsection{MDS Constraints}
Constraints help to define a summary's structure, and the
performance guarantee of the greedy algorithm depends on them. In MDS,
typical constraints are upper bounds in the number of sentences or
words, corresponding to cardinality ($\left|S\right| \le m$) or
knapsack constraints ($\sum_{s \in S} \left|\text{words}(s)\right| \le
m$) for some upper bound $m$. When optimizing a submodular monotone
function under such constraints, \textsc{Greedy} has a performance
guarantee of $\approx 0.63$ and $\approx 0.39$ respectively
\citep{calinescu11,linhui11}. That is, for cardinality constraints,
the output is at least $0.63$ as good as the optimal solution in terms
of objective function value.

\subsection{MDS Objective Functions}
\label{subsec:admds-obj}

In MDS, approaches typically try to maximize coverage and diversity. In its simplest form, \citet{linhui11} model coverage as 
\begin{equation}
  \label{eqn:f_cov}
  f_\text{Cov}(S) = \sum_{s \in S} \sum_{v \in U} \text{sim}(s, v),
\end{equation}
where $\text{sim}\colon\, U \times U \to \mathbb{R}_{\ge 0}$ is a sentence
similarity function, e.g.\ cosine  of word vectors.

\citet{linhui11} model diversity via
\begin{equation}
  \label{eqn:f_div}
  f_\text{Div}(S) = \sum_{i=1}^k \sqrt{\sum_{s \in P_i \cap S} r(s)}
\end{equation}
where $P_1,\ldots,P_k$ is a partition of $U$ (e.g.\ obtained by semantic clustering) and $r\colon\, U \to \mathbb{R}_{\ge 0}$ is a singleton reward function. We get diminished reward for adding additional sentences from one cluster.



\subsection{Application to TLS}
\label{subsec:asmds-appl}

Applying this MDS model to TLS as-is may not be adequate. 
For example, since the length constraints only limit the total number
of sentences, some days in the timeline could be
overrepresented. Furthermore, if objective functions ignore temporal
information, we may not be able to extract sentences that describe
very important events lasting only for short time periods.  Instead,
natural units for TLS are both the whole timeline
as well as  individual dates, so criteria and constraints for TLS should
accommodate both units.


\section{Temporalizing Objective Functions}
\label{sec:obj}

We now systematically add temporal information to the objective
function by (i) temporalizing coverage functions, (ii) temporalizing
diversity functions, and (iii) adding date selection functions.
We prove the monotonicity and submodularity of all functions in the supplementary material.

\subsection{Temporalizing Coverage}

MDS coverage functions  (Equation~\ref{eqn:f_cov}) ignore
temporal information, computing coverage on a
corpus-wide level.  We temporalize
them by modifying the similarity computation.  This is a
minimal but fundamental modification. Previous work in TLS noted that
coverage for candidate summaries for a day $d$ should
look mainly at the temporally local neighborhood, i.e.\ at sentences
whose dates are close to $d$ \citep{chieu04,yanrui11a}.  We
investigate two variants of this idea.
The first uses a hard cutoff \citep{chieu04}, restricting similarity computations to sentences that are at most $p$ days apart:
\begin{equation}
    \text{sim}_p(s, t) = \begin{cases}
        \text{sim}(s, t) & \left|d(s) - d(t)\right| \le p\\
        0 & \left|d(s) - d(t)\right| > p
    \end{cases}
\end{equation}
The second uses a soft variant \citep{yanrui11a}. Let
$g\colon\,\mathbb{N} \to \mathbb{R}_{> 0}$ be monotone with
$g(0) = 1$. We set $\text{sim}^{g}(s, t) = \text{sim}(s,
t)/g(\left|d(s) - d(t)\right|)$.  Thus, all date
differences are penalized, and greater date differences are penalized
more.

\subsection{Temporalizing Diversity}

As with coverage, standard MDS diversity functions 
(Equation \ref{eqn:f_div}) ignore 
temporal information. If the singleton reward $r$  in
$f_\text{Div}$ relies on $\text{sim}$, as is the case with many
implementations, then temporalizing $\text{sim}$  implicitly
temporalizes diversity. We now go beyond such an implicit temporalization.

In TLS, we want to apply diversity on a temporal basis: we do not want
to concentrate the summary on very few, albeit important dates, but we
want date (and subevent) diversity. $f_\text{Div}$, however, typically uses only a semantic
criterion to obtain a partition, e.g.\ by
k-means clustering of sentence vector representations
\citep{linhui11}. This may wrongly conflate  events, such as two unrelated 
protests on different dates.
We can instead employ a
temporal partition. The simplest method is to
partition the sentences by their date, i.e.\ for a temporalized
diversity function $f_\text{TempDiv}$ we have the same form as in
Equation \ref{eqn:f_div}, but $P_i$ contains all sentences with date
$d_i$, where $d_1,\ldots,d_k$ are all sentence dates.


\subsection{Date Selection Criteria}

An important part of TLS is \emph{date selection}. Dedicated
algorithms for date selection use frequency and patterns in date
referencing to determine date importance \citep{tran15a}.
Most  date importance measures  can be integrated into
the objective function to allow for joint date selection and summary
generation.\footnote{Our framework can also be extended to accommodate
  pipelined date selection. We leave this to future
  work.} One well-performing date selection baseline is
to measure for each date how many sentences refer to it. This
objective can be described by the monotone submodular
function 
\begin{equation*}
  f_\text{DateRef}(S) = \sum_{d \in d(S)} \left|\{u \in U ~ \vert ~ u \text{ refers to } d\}\right|.
\end{equation*}

\subsection{Combining Criteria}

We combine coverage, diversity and date importance via unweighted sums
for our final objective functions.  An alternative would be to combine
them via weighted sums learned from training data
\cite{linhui11,linhui12} but since there are only few datasets available for training and  testing
TLS algorithms we choose the unweighted sum to estimate as few parameters as possible from data.


\section{Temporalizing Constraints}
\label{sec:constraints}

The MDS knapsack/cardinality constraints are too simple for TLS as an
overall sentence limit does not constrain a timeline to have daily
summaries of roughly similar length or enforce other uniformity properties. We
introduce constraints going beyond simple cardinality, and prove
performance guarantees of  \textsc{Greedy} under such constraints.

\subsection{Definition of Constraints}

Typically, we have two requirements on the timeline: the total number of days should not exceed a given number $\ell$ and the length of the daily summary (in sentences) should not exceed a given number $k$ (for every day). Let $d$ be the function that assigns each sentence its date. For a set $S \subseteq U$, the requirements can be formalized as
\begin{equation}
  \label{eqn:tl_length}
  \left| \{ d(s) ~ \vert ~ s \in S\} \right| \le \ell
\end{equation}
and, for all $s \in S$,
\begin{equation}
  \label{eqn:daily_length}
  \left| \{ s^\prime ~ \vert ~ s^\prime \in S, ~ d(s^\prime) = d(s)\} \right| \le k.
\end{equation}

\subsection{Performance Guarantees}

While the constraints expressed by Equations \ref{eqn:tl_length} and \ref{eqn:daily_length} are more complex than constraints used in MDS, they have a property in common: if a set $S$ fulfills the constraints (i.e.\ $S \in \mathcal{I}$), then also any subset $T \subseteq S$ fulfills the constraints (i.e.\ $T \in \mathcal{I}$). In combinatorics, such constraints are called \emph{independence systems} \citep{calinescu11}.
\begin{defn}
  Let $V$ be some set and $\mathcal{I} \subset 2^V$ be a collection of subsets of $V$. The tuple $(V, \mathcal{I})$ is called an \emph{independence system} if 
  (i) $\emptyset \in \mathcal{I}$ and (ii) $B \in \mathcal{I}$ and $A \subseteq B$ implies $A \in \mathcal{I}$.
\end{defn}
Optimization theory shows that \textsc{Greedy} also has
performance guarantees when generalizing cardinality/knapsack
constraints to ``sufficiently nice'' independence systems. Based on
these results, we prove Lemma~\ref{lemma::separateconstraints} (see the
suppl.\ material):

\begin{lemma}
  \label{lemma::separateconstraints}
  Let $\mathcal{I}$ be the set of subsets of $U$ that fulfill Equations \ref{eqn:tl_length} and \ref{eqn:daily_length}. Then \textsc{Greedy} has a performance guarantee of $1/(k+1)$.
\end{lemma}

The lemma implies that for small $k$ that is typical in TLS (e.g.\ $k
= 2$), we obtain a good approximation with reasonable constraints.
However, our performance guarantees are still weaker than for MDS (for
example, 0.33 for $k=2$ compared to 0.63 in MDS). The reason for this
is that our constraints are more complex, going beyond the simple
well-studied cardinality and knapsack constraints. We also observe that this is a
worst-case bound: in practice the performance of the algorithm may
approach the exact solution (as \citet{linhui10} show for
MDS). However, such an analysis is out of scope for our paper, since
computing the exact solution is intractable in TLS.\footnote{\citet{mcdonald07} and \citet{linhui10}
already report scalability issues for obtaining
exact solutions for MDS, which is of smaller scale and has simpler
constraints than our task.}


\section{Experiments}
\label{sec::exps}

We evaluate the performance of modeling TLS as MDS and the effect of various temporalizations.

\subsection{Data and Preprocessing}

We run experiments on \emph{timeline17} \citep{tran13a}
and \emph{crisis} \citep{tran15b}. Both data sets consist of
(i) journalist-generated timelines on events such as the Syrian War
as well as (ii) corresponding corpora of news articles
on the topic scraped via Google News. They are publically
available\footnote{\url{http://www.l3s.de/~gtran/timeline/}} and have
been used in previous work \citep{wangwilliam16}.\footnote{The
  datasets used in \citet{chieu04} or \citet{nguyenkiem14} are
  not available.} 
Table \ref{tab:data} shows an overview.


\begin{table}[t]
    \begin{center}
        \footnotesize
        \begin{tabular}{@{}lrrrrr@{}}
            \toprule
            \textbf{Name} & \textbf{Topics} & \textbf{TLs} & \textbf{Docs} & \multicolumn{2}{c}{\textbf{Sentences}}\\
            & & & & \textbf{Total} & \textbf{Filtered}\\
            \midrule
            timeline17 & 9 & 19 & 4,622 & 273,432 & 56,449\\
              crisis & 4 & 22 & 18,246 & 689,165 & 121,803\\
            \bottomrule
        \end{tabular}
    \end{center}
    \caption{\label{tab:data}Data set statistics.}
\end{table}

\begin{table}[t]
    \begin{center}
        \footnotesize
        \begin{tabular}{@{}lllrr@{}}
            \toprule
            \textbf{No} & \textbf{Start} & \textbf{End} & \textbf{Dates} & \textbf{Avg.\ Daily}\\
            & & & & \textbf{Summ. Length}\\
            \midrule
            1 & 2010-04-20 & 2010-05-02 & 13 & 4\\
            2 & 2010-04-20 & 2012-11-15 & 16 & 2\\
            3 & 2010-04-20 & 2010-10-15 & 12 & 2\\
            4 & 2010-04-20 & 2010-09-19 & 48 & 2\\
            5 & 2010-04-20 & 2011-01-06 & 102 & 3\\
            \bottomrule
        \end{tabular}
    \end{center}
    \caption{\label{tab:variation}Properties for the \emph{BP oil spill} timelines in \emph{timeline17}. The corpus contains documents for 218 dates from 2010-04-01 to 2011-01-31.}
\end{table}

In the data sets, even timelines for the same topic have considerable
variation. Table~\ref{tab:variation} shows properties for the
five \emph{BP oil spill} timelines in \emph{timeline17}. There
is substantial variation in range, granularity and average daily summary length.

Following previous work \citep{chieu04,yanrui11a}, we filter
sentences in the corpus using keywords. For each
topic we manually define a set of keywords. If any of the keywords appears in a
sentence, the sentence is retained.

We identify temporal expressions with HeidelTime
\citep{stroetgen13}. If a sentence $s$ contains a time expression that
can be mapped to a day $d$ via HeidelTime we set the date of $s$ to
$d$ (if there are multiple expressions we take the first one).
Otherwise, we set the date of $s$ to the publication date of the
article which contains $s$.\footnote{This procedure is in line with
  previous TLS work \citep{chieu04}. The focus of the current paper is not on
  further improving date
  assignment.}

\subsection{Evaluation Metrics}

Automatic evaluation of TLS is done by ROUGE
\citep{linchinyew04}. We report ROUGE-1 and ROUGE-2 F$_1$ scores for the
\emph{concat}, \emph{agreement} and \emph{align+ m:1} metrics for TLS
we presented in \citet{martschat17}.
These metrics perform
evaluation by concatenating all daily summaries, evaluating only
matching days and evaluating aligned dates based on date and
content similarity, respectively. We evaluate date
selection using F$_1$ score.

\subsection{Experimental Settings}

TLS has no established settings.  Ideally, reference and
predicted timelines should be given the same compression parameters,
such as overall length or number of days.\footnote{This would
  mirror settings in  MDS, where reference and predicted summary have
  the same length constraint.}
  Since there is
considerable variation in timeline parameters (Table
\ref{tab:variation}), we evaluate against each reference timeline individually,
providing  systems with the parameters they need via extraction from the
reference timeline, including range and needed length constraints. We
set $m$ to the number of sentences in the reference timeline, $\ell$
to the number of dates in the timeline, and $k$ to the average length
of the daily summaries.

Most previous work uses different or unreported settings, which makes
comparison difficult. For instance, \citet{tran13a} do not report how
they obtain timeline length. \citet{wanglu15,wangwilliam16} create a constant-length summary for each
day that has an article in the corpus, thereby comparing  reference
timelines with few days 
with predicted timelines that have summaries for each day. 

\subsection{Baselines}

Past work on \emph{crisis} generated summaries from headlines
\cite{wangwilliam16} or only used manual evaluation
\cite{tran15b}. Past work on \emph{timeline17} evaluates with ROUGE
\citep{tran13a,wangwilliam16} but suffers from the fact that
parameters for presented systems, baselines and reference timelines
differ or are not reported (see above). Therefore, we reimplement two baselines that were competitive in
previous work \citep{yanrui11a,wanglu15,wangwilliam16}.  

\paragraph{Chieu.} Our first baseline is
\textsc{Chieu}, the unsupervised approach of \citet{chieu04}. It operates in two stages. First, it ranks sentences based on similarity: for each sentence $s$, similarities to all sentences in a 10-day window around the date of $s$ are summed up\footnote{This corresponds to the \emph{Interest} ranking proposed by \citet{chieu04}. We do not use the more complex \emph{Burstiness} measure since \emph{Interest} was found to perform at least as well in previous work when evaluated with ROUGE-based measures \citep[p.c.]{wanglu15}}. This yields a ranked list of sentences, sorted by highest to lowest summed up similarities. Using this list, a timeline containing one-sentence daily summaries is constructed as follows: iterating through the ranked sentence list, a sentence is added to the timeline depending on the \emph{extent} of the sentences already in the timeline. Extent of a sentence $s$ is defined as the smallest window of days such that the total similarity of $s$ to sentences in this window reaches at least 80\% of the similarity to the sentences in the full 10-day window. If the candidate sentence does not fall into the extent of any sentence already in the timeline, it is added to the timeline.

As we can see, the model and parameters such as daily summary length are intertwined in this approach. We therefore reimplement \textsc{Chieu} exactly instead of giving it reference timeline parameters. As we describe below, we use the same sentence similarity function as \citet{chieu04}.

\paragraph{Regression.} Our second baseline is \textsc{Reg}, a
supervised linear regression model \citep{tran13a,wanglu15}. We represent each sentence with features describing its length, number of named entities, unigram features, and averaged/summed tf-idf scores. During training, for each sentence, standard ROUGE-1 F$_1$ w.r.t.\ the reference summary of the sentence's date is computed. The model is trained to predict this score.\footnote{We use per-topic
  cross-validation \citep{tran13a}.} During prediction, sentences are selected greedily according to predicted F$_1$ score, respecting temporal constraints defined by
the reference timeline.
\subsection{Model Parameters}

For all submodular models and for \textsc{Chieu} we use sparse
inverse-date-frequency sentence representations
\citep{chieu04}\footnote{In preliminary experiments, results using
  such sparse representations were higher than results using dense
  vectors.}. This yields a vector representation $v_s$ for each
sentence $s$. We set $\text{sim}(s, t) = \cos (v_s, v_t)$. We did not
tune any further parameters but re-used settings from previous
work. For modifications to $\text{sim}$  when temporalizing
coverage and diversity (Section~\ref{sec:obj}), we use a cutoff of 10
(as \citet{chieu04}), and consider $g(x) = \sqrt{x+1}$ for
reweighting. We choose the square root since it quickly provides
strong penalizations for date differences but then
saturates. Following \citet{linhui11}, we set singleton reward for
$f_\text{Div}$ to $r(s) = \sum_{u \in U} \text{sim}(s, u)$ and obtain
the partition $P_1,\ldots,P_k$ by k-means clustering with $k = 0.2
\cdot \left|U\right|$. We obtain a temporalization $f_\text{TempDiv}$
of diversity by considering a partition of sentences induced by their
dates (see Section~\ref{sec:obj}).

\subsection{Results}

Results  are displayed
in Table \ref{tab:results}.
The numbers are averaged over all timelines
in the respective corpus. We test for significant differences using an
approximate randomization test \citep{noreen89} with a $p$-value of
$0.05$.

\begin{table*}[ht]
    \begin{center}
        \footnotesize
        \begin{adjustbox}{max width=\textwidth}
        \begin{tabular}{@{}llllllll@{}}
            \toprule
             & \multicolumn{2}{c}{\textbf{concat}} & \multicolumn{2}{c}{\textbf{agree}} & \multicolumn{2}{c}{\textbf{align+ m:1}} & \textbf{Date Sel.}\\
            \textbf{Model} & R1 & R2 & R1 & R2 & R1 & R2 & F$_1$\\
            \midrule
            \multicolumn{8}{c}{\textbf{timeline17}}\\
            \midrule
            Baselines\\
            \hspace{0.5cm}\textsc{Chieu} & 0.296 & 0.072 & 0.039 & 0.016 & 0.066 & 0.019 & 0.251\\
            \hspace{0.5cm}\textsc{Reg} & 0.336 & 0.065 & 0.063 & 0.014 & 0.074 & 0.016 & 0.491\\
            Non-temporal Submodular Models\\
            \hspace{0.5cm}\textsc{AsMDS} & 0.351$^\dag$ & 0.088$^*$ & 0.071$^\dag$ & 0.019 & 0.086$^\dag$ & 0.022 & 0.452$^\dag$\\
            Temporalizing Constraints\\
              \hspace{0.5cm}\textsc{TLSConstraints} & 0.368$^\dag$ & 0.090$^{\dag*}$ & 0.082$^{\dag*}$ & 0.022 & 0.098$^{\dag*}$ & 0.025$^*$ & 0.482$^\dag$\\            
            Temporalizing Criteria\\
              \hspace{0.5cm}\textsc{AsMDS}+cutoff & 0.338$^x$ & 0.083$^*$ & 0.065$^\dag$ & 0.021 & 0.077 & 0.024 & 0.393$^{\dag*x}$\\
              \hspace{0.5cm}\textsc{AsMDS}+reweighting & 0.329$^x$ & 0.081$^x$ & 0.063$^\dag$ & 0.019 & 0.075$^x$ & 0.022 & 0.390$^{\dag*x}$\\
              \hspace{0.5cm}\textsc{AsMDS}+$f_\text{DateRef}$ & 0.357$^\dag$ & \textbf{0.092}$^{\dag*x}$ & 0.082$^{\dag*x}$ & 0.022$^*$ & 0.095$^{\dag*x}$ & 0.025$^*$ & 0.529$^{\dag{x}}$\\
              \hspace{0.5cm}\textsc{AsMDS}+$f_\text{TempDiv}$ & 0.347 & 0.088$^*$ & 0.088$^{\dag*x}$ & 0.026$^{\dag*}$ & 0.103$^{\dag*x}$ & 0.029$^{\dag*x}$ & 0.526$^{\dag{x}}$\\
              \hspace{0.5cm}\textsc{AsMDS}+$f_\text{TempDiv}$+$f_\text{DateRef}$ & 0.347 & 0.090$^*$ & \textbf{0.092}$^{\dag*x}$ & \textbf{0.027}$^{\dag*x}$ & 0.105$^{\dag*x}$ & \textbf{0.030}$^{\dag*x}$ & \textbf{0.544}$^{\dag*x}$\\
            Temporalizing Constraints and Criteria\\
              \hspace{0.5cm}\textsc{TLSConstraints}+cutoff & 0.366$^\dag$ & 0.085$^*$ & 0.091$^{\dag*x}$ & 0.023$^*$ & 0.105$^{\dag*x}$ & 0.026$^*$ & 0.505$^{\dag{x}}$\\
              \hspace{0.5cm}\textsc{TLSConstraints}+reweighting & \textbf{0.371}$^\dag$ & 0.088$^{\dag*}$ & 0.091$^{\dag*x}$ & 0.026$^{\dag*x}$ & \textbf{0.106}$^{\dag*x}$ & 0.028$^{\dag*x}$ & 0.506$^{\dag{x}}$\\
              \hspace{0.5cm}\textsc{TLSConstraints}+$f_\text{DateRef}$ & \textbf{0.371}$^{\dag*x}$ & 0.090$^{\dag*}$ & 0.089$^{\dag*x}$ & 0.023$^*$ & 0.103$^{\dag*x}$ & 0.026$^*$ & 0.517$^{\dag{x}}$\\              
              \hspace{0.5cm}\textsc{TLSConstraints}+$f_\text{DateRef}$+reweighting & 0.370$^{\dag*}$ & 0.091$^{\dag*}$ & 0.090$^{\dag*x}$ & 0.024$^*$ & 0.104$^{\dag*x}$ & 0.027$^*$ & 0.515$^{\dag{x}}$\\
            \midrule
            \multicolumn{8}{c}{\textbf{crisis}}\\
            \midrule
            Baselines\\
            \hspace{0.5cm}\textsc{Chieu} & \textbf{0.374} & 0.070 & 0.029 & 0.008 & 0.052 & 0.012 & 0.142\\
            \hspace{0.5cm}\textsc{Reg} & 0.271 & 0.034 & 0.014 & 0.001 & 0.028 & 0.003 & 0.189\\
            Non-temporal Submodular Models\\
            \hspace{0.5cm}\textsc{AsMDS} & 0.309$^{\dag*}$ & 0.064$^{*}$ & 0.037$^{*}$ & 0.009$^{*}$ & 0.060$^{*}$ & 0.014$^{*}$ & 0.183$^{\dag}$\\
            Temporalizing Constraints\\
              \hspace{0.5cm}\textsc{TLSConstraints} & 0.339$^{\dag*x}$ & 0.066$^{*}$ & 0.035$^{*}$ & 0.008$^{*}$ & 0.058$^{*}$ & 0.012$^{*}$ & 0.180$^{\dag}$\\
            Temporalizing Criteria\\
              \hspace{0.5cm}\textsc{AsMDS}+cutoff & 0.283$^{\dag{x}}$ & 0.061$^{\dag*}$ & 0.036$^{*}$ & 0.011$^{*}$ & 0.050$^{*}$ & 0.014$^{*}$ & 0.186\\
              \hspace{0.5cm}\textsc{AsMDS}+reweighting & 0.294$^{\dag*}$ & 0.061$^{\dag*}$ & 0.039$^{*}$ & 0.011$^{*}$ & 0.056$^{*}$ & 0.015$^{*}$ & 0.212$^{\dag*}$\\
              \hspace{0.5cm}\textsc{AsMDS}+$f_\text{DateRef}$ & 0.314$^{\dag*}$ & 0.067$^{*}$ & 0.042$^{\dag*}$ & 0.009$^{*}$ & 0.065$^{\dag*x}$ & 0.014$^{*}$ & 0.248$^{\dag*x}$\\
              \hspace{0.5cm}\textsc{AsMDS}+$f_\text{TempDiv}$ & 0.311$^{\dag}$ & 0.062$^{*}$ & 0.034$^{*}$ & 0.007$^{*}$ & 0.058$^{*}$ & 0.012$^{*x}$ & 0.196$^{\dag*}$\\
              \hspace{0.5cm}\textsc{AsMDS}+$f_\text{TempDiv}$+$f_\text{DateRef}$ & 0.311$^{\dag*}$ & 0.064$^{*}$ & 0.039$^{\dag*}$ & 0.008$^{*}$ & 0.063$^{\dag*}$ & 0.012$^{*}$ & 0.233$^{\dag*x}$\\
            Temporalizing Constraints and Criteria\\
              \hspace{0.5cm}\textsc{TLSConstraints}+cutoff & 0.323$^{\dag*x}$ & 0.068$^{*}$ & 0.046$^{\dag*}$ & 0.011$^{*}$ & 0.066$^{\dag*}$ & 0.015$^{*}$ & 0.242$^{\dag{x}}$\\
              \hspace{0.5cm}\textsc{TLSConstraints}+reweighting & 0.332$^{\dag*x}$ & 0.071$^{*x}$ & 0.044$^{\dag*}$ & 0.009$^{*}$ & 0.068$^{\dag*}$ & 0.014$^{*}$ & 0.237$^{\dag{x}}$\\
              \hspace{0.5cm}\textsc{TLSConstraints}+$f_\text{DateRef}$ & 0.333$^{\dag*x}$ & 0.069$^{*x}$ & 0.045$^{\dag*x}$ & 0.009$^{*}$ & 0.067$^{\dag*x}$ & 0.013$^{*}$ & 0.248$^{\dag*x}$\\
              \hspace{0.5cm}\textsc{TLSConstraints}+$f_\text{DateRef}$+reweighting & 0.333$^{\dag*x}$ & \textbf{0.072}$^{*x}$ & \textbf{0.054}$^{\dag*x}$ & \textbf{0.012}$^{\dag*}$ & \textbf{0.075}$^{\dag*x}$ & \textbf{0.016}$^{*}$ & \textbf{0.281}$^{\dag*x}$\\
            \bottomrule
        \end{tabular}
        \end{adjustbox}
    \end{center}
    \caption{\label{tab:results}Results. Highest values per column/dataset are boldfaced. For the submodular models, $^\dag$ denotes sign.\ difference to \textsc{Chieu}, $^*$ to \textsc{Reg}, $^x$ to \textsc{AsMDS}.}
\end{table*}

\paragraph{Baselines.}

Overall, performance on
\emph{crisis} is much lower than on \emph{timeline17}. This is because (i) 
the corpora in \emph{crisis} contain
articles for more days over a larger time span and (ii) 
average percentage of
article publication dates for which a summary in a corresponding reference timeline exists is $11\%$ for
\emph{timeline17} and $3\%$ for \emph{crisis}. This makes date
selection more difficult.
On \emph{crisis}, \textsc{Chieu} outperforms \textsc{Reg} except
for date selection. On \emph{timeline17}, \textsc{Reg}
outperforms \textsc{Chieu} for four out of seven metrics.
Timelines in \emph{crisis}  contain
fewer dates and shorter daily summaries than timelines in
\emph{timeline17}, which aligns well with \textsc{Chieu}'s redundancy
post-processing.

\paragraph{TLS as MDS.}
The model \textsc{AsMDS} uses standard length constraints from MDS
and an objective function combining non-temporalized $f_\text{Cov}$
and $f_\text{Div}$. It allows us to evaluate how well standard MDS ports
to TLS.
Except for \emph{concat} and \emph{date selection} on \emph{crisis}, this model outperforms
both baselines, while providing the advantages of modularity,
non-supervision and feature/inference separation discussed throughout
the paper.

\begin{table*}[t]
    \begin{center}
        \footnotesize
        \begin{adjustbox}{max width=\textwidth}
        \begin{tabular}{@{}lrrr|rrr@{}}
            \toprule
            \textbf{Name} & \multicolumn{3}{c}{\textbf{Compression rate} $r$} & \multicolumn{3}{c}{\textbf{Spread} $s$}\\
            & $r \in [0, 0.001]$ & $r \in (0.001,0.01]$ & $r \in (0.01,0.1]$ & $s \in [0,1/3]$ & $s \in (1/3,2/3]$ & $s \in (2/3,1]$\\
            \midrule
            \textsc{Chieu} & 0.06 & 0.08 & 0.07 & 0.06 & 0.08 & 0.04\\
            \textsc{Reg} & 0.04 & 0.09 & 0.07 & 0.05 & 0.11 & 0.11\\
            \textsc{AsMDS} & 0.05 & 0.10 & 0.09 & 0.07 & 0.10 & 0.10\\
            \textsc{TLSConstraints}  & 0.08 & 0.10 & 0.10 & 0.08 & 0.12 & 0.14\\
            \textsc{AsMDS}+$f_\text{TempDiv}$+$f_\text{DateRef}$ & 0.09 & 0.11 & 0.12 & 0.09 & 0.13 & 0.13\\
            \bottomrule
        \end{tabular}
        \end{adjustbox}
    \end{center}
    \caption{\label{tab:analysis_compr}Results (\emph{align+ m:1} ROUGE-1 F$_1$) by \emph{compression rate} and \emph{spread} on timeline17.}
\end{table*}
\begin{table}[t]
    \begin{center}
        \footnotesize
        \begin{tabular}{@{}lrr@{}}
            \toprule
            \textbf{Name} & \multicolumn{2}{c}{\textbf{Max.\ Length}}\\
            & Mean & Median \\
            \midrule
            Reference & 5.6 $\pm$ 2.7 & 5\\
            \midrule
            \textsc{Chieu} & 1.0 $\pm$ 0.0 & 1\\
            \textsc{Regression} & 2.3 $\pm$ 1.7 & 2\\
            \textsc{AsMDS} & 23.7 $\pm$ 41.2 & 8\\
            \textsc{TLSConstraints} & 2.3 $\pm$ 1.7 & 2\\
            \textsc{AsMDS}+$f_\text{TempDiv}$+$f_\text{DateRef}$ & 3.8 $\pm$ 5.3 & 1\\
            \bottomrule
        \end{tabular}
    \end{center}
    \caption{\label{tab:analysis_length_and_red}Length of longest daily summary, mean and median over all timelines on \emph{timeline18}.}
\end{table}
\begin{table}[ht]
    \begin{center}
        \footnotesize
        \begin{tabular}{@{}lrrrrrrr@{}}
            \toprule
             & \multicolumn{2}{c}{\textbf{concat}} & \multicolumn{2}{c}{\textbf{agree}} & \multicolumn{2}{c}{\textbf{align+ m:1}} & \textbf{Date}\\
            \textbf{Corpus} & R1 & R2 & R1 & R2 & R1 & R2 & F$_1$\\
            \midrule
            tl17 & 0.50 & 0.18 & 0.30 & 0.14 & 0.30 & 0.14 & 0.87\\
            crisis & 0.49 & 0.16 & 0.34 & 0.14 & 0.35 & 0.14 & 0.95\\
            \bottomrule
        \end{tabular}
    \end{center}
    \caption{\label{tab:oracle}Oracle results optimizing per-day R1 F$_1$.}
\end{table}

\paragraph{Temporalizing Constraints.}
The model \textsc{TLSConstraints} uses the temporal constraints described in
Section \ref{sec:constraints}, but has the same objective function as \textsc{AsMDS}.  Compared to \textsc{AsMDS}, there are
improvements on all metrics on \emph{timeline17} and 
similar performance on \emph{crisis}.

\paragraph{Temporalizing Criteria.}
We temporalize \textsc{AsMDS} objective functions
(Section~\ref{sec:obj}) via modifications of the similarity function
(cutoffs/reweightings), replacing diversity by temporal
diversity $f_\text{TempDiv}$, and adding date selection
$f_\text{DateRef}$.  Constraints are kept non-temporal. If modifications improve over \textsc{AsMDS} we
also check for cumulative improvements. Modifying similarity is not
effective, results drop or stay roughly the same according to most
metrics. The other modifications improve performance w.r.t.\ most
metrics, especially for date selection.

\paragraph{Temporalizing Constraints and Criteria.}
Lastly, we evaluate the joint contribution of temporalized constraints
and criteria.\footnote{We do not evaluate $f_\text{TempDiv}$, since the
temporal constraints already capture temporal diversity.}
Modifications to the similarity
function have a positive effect, especially reweighting. $f_\text{DateRef}$
provides information about date importance not encoded in the
constraints, improving results on \emph{crisis}.


\paragraph{Oracle Results.}
Previous research in MDS computed oracle upper bounds (e.g. \citet{hirao17}). To estimate TLS
difficulty and our limitations, we provide the first oracle upper
bound for TLS: For each sentence $s$,
we compute ROUGE-1 F$_1$ $g_s$ w.r.t.\ the reference summary for the
sentence's date. We then run \textsc{Greedy} for $f_\text{Oracle}(S) =
\sum_{s \in S} g_s$, employing the same constraints as
\textsc{TLSConstraints} (see Table \ref{tab:oracle}).

Scores of the models are most similar to oracle results for
the temporally insensitive \emph{concat} metric, with gaps
comparable to gaps in MDS \citep{hirao17}. The biggest gap is
in \emph{date selection} F$_1$. This also leads to higher
differences in the scores of temporally sensitive metrics,
highlighting the importance of temporal
information.

\subsection{Analysis}

We now investigate where and how
temporal information helps compared to \textsc{AsMDS}. We have already identified two potential weaknesses of
modeling TLS as MDS: the low compression rate (Section
\ref{sec:task}) and the likely case that \textsc{AsMDS} overrepresents
certain dates in a timeline (Section \ref{sec:submod}). We now
analyze the behavior of \text{AsMDS} w.r.t.\ these points and
discuss the effect of temporal information. To
avoid clutter, we restrict analysis to \emph{timeline17} and
report only \emph{align+ m:1} ROUGE-1 F$_1$.

\paragraph{Effect of Compression Rate.}

We hypothesize that difficulty
increases as compression rate decreases. We measure compression rate in two
ways. We first adopt the definition from MDS and define
\emph{corpus compression rate} as the number of sentences in a reference
timeline divided by the number of sentences in the (unfiltered)
corresponding corpus. Second, we define a TLS-specific
notion called \emph{spread} as the number of dates in the reference timeline divided
by the maximum possible number of dates given its start and end
date. For example, the timeline from Table \ref{table:example} in the
introduction has spread $3/14$. 
We see that timelines with lowest compression rate/spread are indeed
the hardest (Table~\ref{tab:analysis_compr}). Temporal information
leads to improvements in all categories.

\paragraph{(Over)representation of Dates.}
We hypothesized that \textsc{AsMDS} may overrepresent certain
dates. We test this hypothesis by measuring the length (in sentences)
of the longest daily summary in a timeline, and computing mean
and median over all timelines (Table \ref{tab:analysis_length_and_red}).
The numbers confirm the hypothesis: When modeling TLS as MDS, some
daily summaries tend to be very long. By construction of the
constraints employed, the effect does not occur or is much weaker for
\textsc{Chieu}, \textsc{Reg} and
\textsc{TLSConstraints}. Temporal objective functions (as in
\textsc{AsMDS}+$f_\text{TempDiv}$+$f_\text{DateRef}$) also weaken the
effect substantially.


\section{Related Work}
\label{sec:rel}

The earliest work on TLS is \citet{allan01}, who
introduce the concepts of usefulness (conceptually similar to
coverage) and novelty (similar to
diversity), using a simple multiplicative
combination.  However, both concepts are
not temporalized.
The notion of usefulness is developed further as ``interest''
 by \citet{chieu04}, which we use as one of
our baselines. \citet{chieu04} compute interest/coverage in a
static local date-based window, instead of using global optimization
as we do. They handle redundancy only during
post-processing s.t.\ the interplay between coverage and diversity is not
adequately modeled.  Further optimization criteria are introduced by
\citet{yanrui11a,yanrui11b} and \citet{nguyenkiem14}, but their frameworks suffer from a lack of
modularity or from an unclear separation of features and architecture.
\citet{wanglu15}  devise a local submodular model for predicting daily
summaries in TLS, but they do not model the whole timeline generation as
submodular function optimization under suitable constraints.

\citet{wangwilliam16} tackle only the task of generating daily
summaries without date selection using a supervised framework,
greedily optimizing per-day predicted ROUGE scores, using images
and text. In contrast, \citet{kessler12} and \citet{tran15a} only tackle date
selection but do not generate any summaries. We
consider the full task, including date selection and summary generation.

TLS is related to standard MDS. We discussed differences in Section~\ref{sec:task}.
%
Our framework is inspired by
\citet{linhui11} who  cast MDS as optimization of submodular
functions under cardinality and knapsack constraints. We go beyond
their work by  modeling temporally-sensitive objective functions as well as
 more complex constraints encountered in TLS.

 A related task is TREC \emph{real-time summarization} (RTS) \citep{linjimmy16}.\footnote{Predecessors of this task were the \emph{update} and \emph{temporal summarization tasks} \citep{aslam15}}.  In contrast
 to TLS, this task requires \emph{online} summarization by presenting
 the input as a stream of documents and emphasizes novelty detection
 and lack of latency. In addition, RTS focuses on
 social media and has a very fine-grained temporal granularity.
 TLS also has an emphasis on date selection and dating for algorithms
 and evaluation which is not present in RTS as the
 social media messages are dated a priori.



\section{Conclusions}
\label{sec:conc}

We show that submodular optimization models for MDS
can yield well-performing models for TLS, despite
the differences between the tasks. Therefore we can port advantages
such as modularity and separation between features and inference,
which current TLS models lack. In addition, we temporalize these MDS-based models to take into
account TLS-specific properties, such as timeline uniformity
constraints, importance of date selection and temporally sensitive
 objectives.  These temporalizations increase
performance without losing the mentioned advantages. We prove that the
ensuing functions are still submodular and that the more complex
constraints still retain performance guarantees for a greedy
algorithm, ensuring scalability.




\section*{Acknowledgments}

We thank the anonymous reviewers and our colleague Josef Ruppenhofer for feedback on earlier drafts of this paper.

\bibliography{../../../../bib/martschat.bib}
\bibliographystyle{acl_natbib_nourl}

\end{document}


\maketitle
\begin{abstract}
  Contains proofs.
\end{abstract}

\section{Proofs}
\label{sec:appendix}

\subsection{Proof of Lemma 1}

For more details and the combinatorial background regarding the used lemmas and theorems we refer the reader to \citet{calinescu11}.

\subsection{Performance Guarantees for Independence Systems}

In order to relate independence systems to performance guarantees for the greedy algorithm we need the notion of a \emph{base}.

\begin{defn}
  Let $(V, \mathcal{I})$ be an independence system. Let $X \subseteq V$. The set of bases of $X$ is
  \begin{equation}
    \mathcal{B}(X) = \{A ~ \vert ~ A \subseteq X, A \in \mathcal{I} \text{ and } A \text{ is maximal}\}
  \end{equation}
\end{defn}

In general a set can have multiple bases. The performance guarantee of the greedy algorithm depends on the relation of the largest to the smallest base.

\begin{defn}
  Let $(V, \mathcal{I})$ be an independence system. Let $X \subseteq V$. We define the \emph{lower rank} of $X$, $\mathrm{lr}(X)$, as the size of the smallest base of $X$. We define the \emph{upper rank} of $X$, $\mathrm{ur}(X)$, as the size of the largest base of $X$.
\end{defn}

\begin{defn}
  Let $p \in \mathbb{N}$. An independence system $(V, \mathcal{I})$ is a \emph{$p$-independence system} if for each $X \subseteq V$ the size of the largest base of $X$ is at most $p$ times the size of the smallest base of $X$, i.e.\
  \begin{equation}
    \frac{\mathrm{ur}(X)}{\mathrm{lr}(X)} \le p
  \end{equation}
\end{defn}

We can now give the main algorithmic result needed for the proof of our lemma \citep{fisher78,calinescu11}. 

\begin{thm}
  \label{thm:main}
  Let $(V, \mathcal{I})$ be a $p$-independence system and let $f\colon\,2^V \to \mathbb{R}$ be a submodular monotone function. Then the greedy algorithm solves the optimization problem
  \begin{equation}  
    \max_{X \subseteq V, X \in \mathcal{I}} f(X)
  \end{equation}
  within the constant factor $1/(p+1)$.
\end{thm}

Now we can prove the lemma.

\begin{lemma}
Let
\begin{equation}
  \label{eqn:tl_length}
  \left| \{ d(s) ~ \vert ~ s \in S\} \right| \le \ell
\end{equation}
and, for all $s \in S$,
\begin{equation}
  \label{eqn:daily_length}
  \left| \{ s^\prime ~ \vert ~ s^\prime \in S, ~ d(s^\prime) = d(s)\} \right| \le k.
\end{equation}
Then the following holds: Let $\mathcal{I}$ be the set of subsets of $U$ that fulfill Equations \ref{eqn:tl_length} and \ref{eqn:daily_length}. Then $\mathcal{I}$ is a $k$-independence system.
\begin{proof}
    Let $Y \subseteq U$. We need to show that
    \begin{equation}
      \frac{\max_{A \in \mathcal{B}(Y)} \left|A\right|}{\min_{A \in \mathcal{B}(Y)} \left|A\right|} \le k.
    \end{equation}
    In order to show this, we show that for
    \begin{equation}
      m = \min\left\{\ell, \left|\{ d(s) ~ \vert ~ s \in Y\}\right|\right\}
    \end{equation}
    it holds that
    \begin{equation}
      \label{eqn:upper}
      \max_{A \in \mathcal{B}(Y)} \left|A\right| \le mk
    \end{equation}
    and
    \begin{equation}
      \label{eqn:lower}
      \min_{A \in \mathcal{B}(Y)} \left|A\right| \ge m.
    \end{equation}    
    Combining Equations \ref{eqn:upper} and \ref{eqn:lower} we arrive at
    \begin{equation}
      \frac{\max_{A \in \mathcal{B}(Y)} \left|A\right|}{\min_{A \in \mathcal{B}(Y)} \left|A\right|} \le \frac{mk}{m} = k
    \end{equation}
    which proves the result.

    \paragraph{Upper Bound.} We first prove the upper bound (Equation \ref{eqn:upper}). Let $A \in \mathcal{B}(Y)$. We consider the equivalence relation $\sim_d$ defined on $A \times A$ by $a \sim_d a^\prime$ if and only if $d(a) = d(a^\prime)$. This equivalence relation induces a partition of $A$ into its equivalence classes according to $\sim_d$. We therefore have
    \begin{align}
      \left|A\right| & = \sum_{[a] \in A/\sim_d} \left|[a]\right|\\
                     & = \sum_{[a] \in A/\sim_d} \left| \{a^\prime \in A ~ \vert ~ d(a^\prime) = d(a)\} \right|\\
                     & \le \sum_{[a] \in A/\sim_d} k\\
                     & = mk.
    \end{align}
    The inequality follows from Equation \ref{eqn:daily_length} and the final equality follows from Equation \ref{eqn:tl_length}, since the equation implies that $\sim_d$ has at most $m$ equivalence classes. Since the estimate holds for every $A \in \mathcal{B}(Y)$, it follows that $\max_{A \in \mathcal{B}(Y)} \left|A\right| \le mk$.

    \paragraph{Lower Bound.} We now prove the lower bound (Equation \ref{eqn:lower}). We use a proof by contradiction. Hence, assume that
    \begin{equation}
      \min_{A \in \mathcal{B}(Y)} \left|A\right| < m.
    \end{equation}
    This implies that there exists $A \in \mathcal{B}(Y)$ with $\left|A\right| < m$. Since $m = \min\left\{\ell, \left|\{ d(s) ~ \vert ~ s \in Y\}\right|\right\}$, there exists $s \in Y$ with $d(s) \notin \{d(s^\prime) ~ \vert ~ s^\prime \in A\}$ and therefore $s \notin A$. Then consider $A^\prime = A \cup \{s\}$. We now show that $A^\prime \in \mathcal{I}$, which is a contradiction to the maximality of $A \in \mathcal{B}(Y)$.

    We first observe that
    \begin{align}
        \left|\{ d(a^\prime) ~ \vert ~ a^\prime \in A^\prime \} \right| & = \left|\{ d(a) ~ \vert ~ a \in A \} \right| + 1\\
                                                                        & < m + 1 \le m \le \ell.
    \end{align}
    
    We still need to show that Equation \ref{eqn:daily_length} holds for all $a^\prime \in A^\prime$. Equation \ref{eqn:daily_length} holds for all $a \in A$ since $d(s) \notin \{d(a) ~ \vert ~ a \in A\}$. For the same reason,
    \begin{equation}
      \left| \{ a^{\prime} ~ \vert ~ a^{\prime} \in A^\prime, ~ d(a^{\prime}) = d(s)\} \right| = \left|\{s\}\right| = 1 \le k.
    \end{equation}        
    Hence, Equation \ref{eqn:daily_length} holds for all $a^\prime \in A^\prime$, which proves the contradiction that $A^\prime \in \mathcal{I}$.
\end{proof}
\end{lemma}

\subsection{Proofs of submodularity and monotonicity of the objective functions}

Submodularity and monotonicity of $f_\text{Cov}$ and $f_\text{Div}$ is shown by \citet{linhui11}. Since our modifications to the similarity functions preserve the non-negativity, the temporalized versions of $f_\text{Cov}$ are also monotone and submodular. $f_\text{TempDiv}$ is monotone and submodular since it has the same form as $f_\text{Div}$. Therefore we just need to show that $f_\text{DateRef}$ is monotone and submodular.

\begin{lemma}
  $f_\text{DateRef}$ is monotone and submodular.
  \begin{proof}
  We first show monotonicity and then submodularity. For brevity, we write $f$ instead of $f_\text{DateRef}$.

  \paragraph{Monotonicity.}  Let $A \subseteq B \subseteq U$. Then
  \begin{equation}
    f(A) = \sum_{d \in d(A)} \left|\{u \in U ~ \vert ~ u \text{ refers to } d\}\right|  
  \end{equation}
  Since $A \subseteq B$ we have $d(A) \subseteq d(B)$ and 
  \begin{align*}
    f(B) = & f(A)\\
                          & + \sum_{d \in d(B) \setminus d(A)} \left|\{u \in U ~ \vert ~ u \text{ refers to } d\}\right|
  \end{align*}
  Since all elements in the sums are non-negative it follows that $f(A) \le f(B)$. Therefore $f$ is monotone.

  \paragraph{Submodularity.} Let $A \subseteq B \subset U$, $v \in U \setminus B$.
  Then
  \begin{equation*}
    f(A \cup \{v\}) - f(A) = \begin{cases}
      f(\{v\}) & d(v) \notin d(A)\\
      0 & \text{otherwise}
    \end{cases}
  \end{equation*}
  and
  \begin{equation*}
    f(B \cup \{v\}) - f(B) = \begin{cases}
      f(\{v\}) & d(v) \notin d(B)\\
      0 & \text{otherwise}
    \end{cases}
  \end{equation*}  
  We need to show that $f(A \cup \{v\}) - f(A) \ge f(B \cup \{v\}) - f(B)$.

  Since $A \subseteq B$, $d(A) \subseteq d(B)$. Now we consider two cases:

  $d(v) \notin d(A)$: Then $f(A \cup \{v\}) - f(A) = f(\{v\})$. If $d(v) \notin d(B)$, then $f(B \cup \{v\}) - f(B) = f(\{v\})$. If $d(v) \in d(B)$, then $f(B \cup \{v\}) - f(B) = 0$. Hence, $f(A \cup \{v\}) - f(A) \ge f(B \cup \{v\}) - f(B)$.

  $d(v) \in d(A)$: Then $d(v) \in d(B)$, and therefore $f(A \cup \{v\}) - f(A) = f(B \cup \{v\}) - f(B) = 0$, hence $f(A \cup \{v\}) - f(A) \ge f(B \cup \{v\}) - f(B)$.
  \end{proof}
\end{lemma}

\bibliography{../../../../bib/martschat.bib}
\bibliographystyle{acl_natbib_nourl}